\documentclass{article}

\usepackage[sort, numbers]{natbib}
\PassOptionsToPackage{numbers, compress}{natbib}

\usepackage{arxiv}

\usepackage[utf8]{inputenc} 
\usepackage[T1]{fontenc}    
\usepackage{hyperref}       
\usepackage{url}            
\usepackage{booktabs}       
\usepackage{amsfonts}       
\usepackage{amsmath}        
\usepackage{graphicx}       
\usepackage{nicefrac}       
\usepackage{microtype}      
\usepackage{xcolor}         
\usepackage{times}
\usepackage{latexsym}
\usepackage{graphicx}
\usepackage{xspace}
\usepackage{helvet}
\usepackage{tcolorbox}
\usepackage{enumitem}
\usepackage{verbatim}
\usepackage{anyfontsize}
\usepackage{amsmath}
\usepackage{amssymb}
\usepackage{wrapfig}
\usepackage{subcaption}
\usepackage{tabu}
\usepackage{algorithm}
\usepackage{multicol}
\usepackage{multirow}
\usepackage{listings}
\usepackage{colortbl}
\usepackage{booktabs}
\usepackage{makecell}
\usepackage{tabularx}
\usepackage{lipsum}
\usepackage{bbm}

\usepackage{hyperref}
\usepackage{xcolor}
\definecolor{citecolor}{HTML}{0071BC}
\hypersetup{colorlinks,linkcolor={red},citecolor={citecolor}}

\newcommand{\methodname}{ReasonManip\xspace}

\title{Incentivizing Multimodal Reasoning in Large Models\\for Direct Robot Manipulation}

\author{%
  Weiliang Tang\textsuperscript{*} \\
  Department of Computer Science \\The Chinese University of Hong Kong \\
\And
  Dong Jing\textsuperscript{*} \\
  Gaoling School of Artificial Intelligence \\Renmin University of China \\ 
  \And
  Jia-Hui Pan \\
  Department of Computer Science\\
  The Chinese University of Hong Kong \\
  \And
  Zhiwu Lu \\
  Gaoling School of Artificial Intelligence\\
  Renmin University of China \\
  \And
  Yun-Hui Liu\\
  Department of Computer Science\\
  The Chinese University of Hong Kong \\
  \And
   Li Erran Li \\
   AWS AI\\ Amazon \\
  \And
   Mingyu Ding \\
   Department of Computer Science\\University of North Carolina at Chapel Hill \\
  \And
  Chi-Wing Fu \\
  Department of Computer Science\\The Chinese University of Hong Kong \\
  \renewcommand\footnotemark{}

  \thanks{
  \textsuperscript{*} Equal contributions to the works. Acknowledgement statement for publications related to HKCLR has been updated: This study was [supported/funded in part] by the InnoHK initiative of the Innovation and Technology Commission of the Hong Kong Special Administrative Region Government via the Hong Kong Centre for Logistics Robotics.}
}

\begin{document}

\maketitle

\begin{abstract}
Recent Large Multimodal Models (LMMs) have demonstrated remarkable reasoning capabilities, especially in solving complex mathematical problems and realizing accurate spatial perception.
Our key insight is that these emerging abilities can naturally extend to robotic manipulation by enabling LMMs to \textbf{directly} infer the next goal (\emph{e.g.} target gripper poses) in language via reasoning, rather than relying on a separate action head.
However, this paradigm meets two main challenges: \textbf{i)} How to make LMMs understand the spatial action space, and \textbf{ii)} How to fully exploit the reasoning capacity of LMMs in solving these tasks.
To tackle the former challenge, we propose a novel task formulation, which inputs the current states of object parts and the gripper, and reformulates rotation by a new axis representation instead of traditional Euler angles.
This representation is more compatible with spatial reasoning and easier to interpret within a unified language space.
For the latter challenge, we design a pipeline to utilize cutting-edge LMMs to generate a small but high-quality reasoning dataset of multi-round dialogues that successfully solve manipulation tasks for supervised fine-tuning.
Then, we perform reinforcement learning by trial-and-error interactions in simulation to further enhance the model's reasoning abilities for robotic manipulation.
Our resulting reasoning model built upon a 7B backbone, named \textbf{\methodname}, demonstrates three notable advantages driven by its system-2 level reasoning capabilities: 
\textbf{i)} exceptional generalizability to out-of-distribution environments, objects, and tasks; 
\textbf{ii)} inherent sim-to-real transfer ability enabled by the unified language representation shared across domains; 
\textbf{iii)} transparent interpretability connecting high-level reasoning and low-level control.
Extensive experiments demonstrate the effectiveness of the proposed paradigm and its potential to advance LMM-driven robotic manipulation.
\end{abstract}

\section{Introduction}

Large Multimodal Models (LMMs) have demonstrated strong and generalizable performance on vision-language understanding and feedback.
Recent efforts have explored leveraging LMMs for robotic manipulation, primarily through two paradigms: textual rule generation and end-to-end action prediction.
%
The first paradigm~\cite{huang2023voxposer,liang2023code,tang2025geomanip} employs LMMs to generate textual rules or codes for restricting robot moving trajectory, which can be integrated with conventional motion controllers. While these rules are interpretable, they often prove difficult for controllers to execute effectively due to the gap between high-level commands and specific robotic actions.
%
The second paradigm~\cite{brohan2022rt, walke2023bridgedata}, also known as Vision-Language-Action (VLA) models, directly maps visual observations and user instructions to executable actions.
Though more flexible, VLA models typically lack deep reasoning capabilities.
Recent LMM-based methods, like SayCan~\cite{ahn2022can}, ECoT~\cite{zawalski2024robotic}, and HiRobot~\cite{shi2025hi}, attempt to combine both paradigms and improve reasoning by generating intermediate perceptual descriptions and planning steps.
However, the action generation in these methods remains implicit, \emph{i.e.,} through action models or separate heads, resulting in a weak connection between reasoning content and the actual actions.
Moreover, these approaches fall short of achieving system-2-level reasoning~\cite{kahneman2011thinking}, which is a deliberate and conscious effort for action generation, i.e., detailed action derivation, environmental check, failure discovery, and task reflection.
Therefore, how to stimulate the deep reasoning capabilities of LMMs and directly apply to robot manipulation remains a key open question.


\begin{figure}
    \centering
    \includegraphics[width=0.95\linewidth,height=0.32\linewidth]{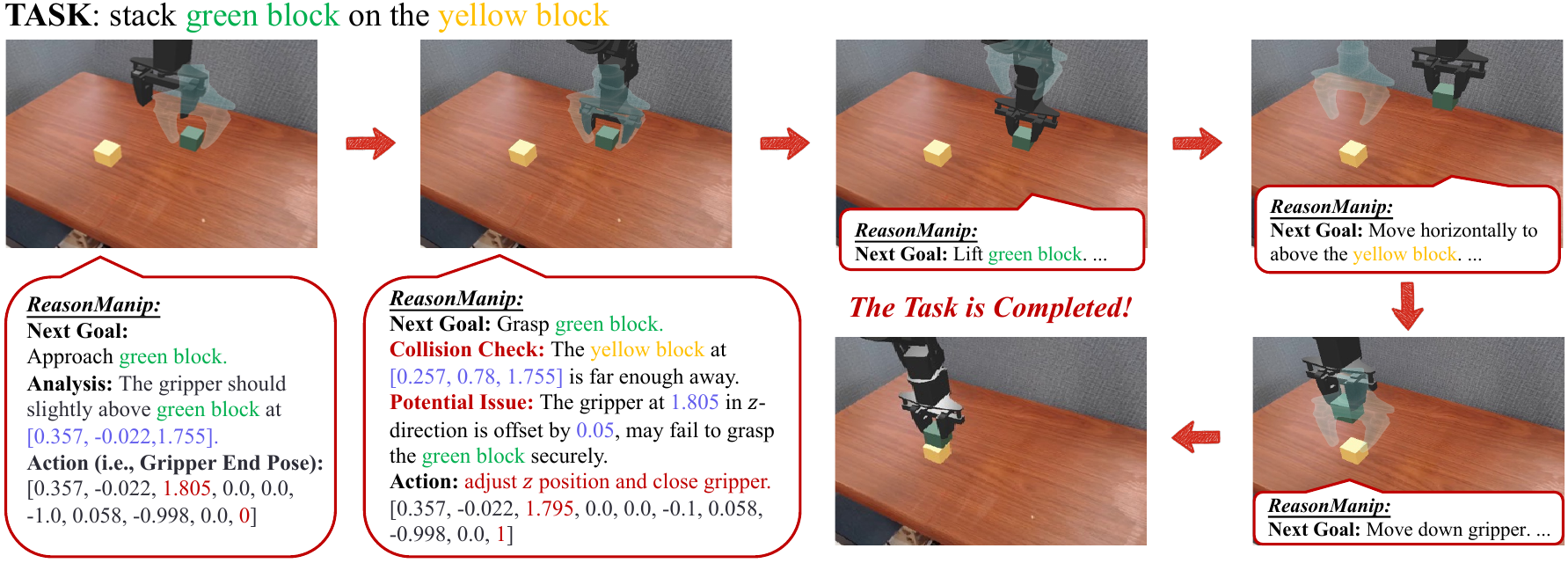}
    \vspace{-4pt}
    \caption{An illustration of \methodname that solves robot manipulation tasks by next-goal prediction via unified system-2 level reasoning, including status infering, collision check, failure mode discovery, task replanning and reflection, etc.
    Unlike prior approaches~\cite{zawalski2024robotic} that rely on separate high-level task decomposition and low-level action models or action heads, \methodname enables LMMs to ``directly'' infer the next goal in natural language (\emph{e.g.} target gripper poses) via multimodal reasoning.}
    \label{fig:intro_paradigm}
    \vspace{-19pt}
\end{figure}

To this end, considering that most robot manipulation tasks could be naturally achieved in a list of next-goal predictions, \emph{e.g.,} gripper end poses. If we can reason about and sequentially predict these intermediate goals, the robot can be guided to reach these targets and complete the task.
Recent advances in the reasoning capabilities of LMMs, particularly in solving complex mathematical problems and spatial reasoning, present new opportunities to apply this \textbf{goal-driven reasoning paradigm} for robotic end pose estimation as a mathematical problem.
This paradigm meets two critical challenges.
\textbf{i)} How to build the connection between visual observations and the spatial action space, enabling LMMs to output precise poses of the gripper's positions and rotation angles?
\textbf{ii)} How to fully leverage the reasoning capabilities of LMM to achieve reliable next goal prediction, spatial perception, and mathematical calculation for solving robot manipulation tasks?


To tackle the above challenges, we first design a novel task formulation as next-goal-wise multi-round dialogues with a new axis-based rotation representation, which is better aligned with LMMs’ reasoning capabilities than traditional Euler angles~\cite{slabaugh1999computing}. Given the quantitative state of object parts and the gripper extracted from point clouds as the input, our new formulation gives LMMs structured access to spatial information and allows them to reason step-by-step through natural language. 
As shown in Fig.~\ref{fig:intro_paradigm}, each round of interaction involves the LMM analyzing the current scene context and producing the next goal as a concrete gripper pose estimate after careful system-2 level reasoning. This transforms the trajectory planning process into a sequence of explicit spatial reasoning and mathematical derivations, naturally aligning with LMMs' capabilities.
Notably, the entire reasoning process encompassing high-level task decomposition, mid-level goal analysis, and low-level pose estimation, are within a single reasoning dialogue pass of the LMM in unified language representation.



Specifically, our training paradigm for building \textbf{\methodname} consists of two stages.
First, we curate a seed dataset of 65 high-quality multi-round dialogues by prompting Qwen2.5-VL-72B-Instruct~\cite{Qwen2.5-VL} to solve tasks step-by-step in the SIMPLER simulator~\cite{li24simpler} with user guidance.
Supervised fine-tuning (SFT) is then applied to this data, resulting in a baseline model with initial capabilities for solving tasks in a multi-round reasoning paradigm.
Next, we perform Group Relative Policy Optimization (GRPO)~\cite{shao2024deepseekmath} to the SFT model by interacting with the SIMPLER simulator to further boost its spatial and mathematical reasoning abilities.
Interestingly, we observe that GRPO not only improves the task success rate, but also but also encourages emergent reasoning and reflection behaviors, signaling the development of system-2 level reasoning.

\methodname achieves strong performance compared with existing VLA approaches,
offering three key advantages based on system-2 level reasoning:
\textbf{(i) High out-of-distribution generalization.} Due to the unified language interface and generalizable spatial representations, \methodname trained in SIMPLER transfers effectively to unseen Meta-World tasks and even real-world robot settings. GRPO plays a significant role in enhancing this generalization.
\textbf{(ii) Robustness to viewpoint changes.} The axis-based spatial representation and reasoning-driven design allow \methodname to remain reliable even under significant variations in camera perspective.
\textbf{(iii) Interpretability from reasoning to actions.} By predicting intermediate goals through explicit dialogue and spatial reasoning, \methodname offers transparency between high-level decision-making and low-level execution.

Overall, our contributions are three-fold:
\textbf{(i)} We reformulate robot manipulation into multi-round mathematical derivation problem with novel axis-based spatial representations, allowing the reasoning abilities of LMMs to directly infer intermediate goals and end-effector poses via language.
\textbf{(ii)} Our user-guided data collection pipeline and two-stage training strategy pave the way for leveraging GRPO to effectively incentivize system-2-level reasoning in LMMs for unifying high-level decision-making and low-level execution.
\textbf{(iii)} We demonstrate through real and simulated experiments that \methodname offers strong data efficiency, out-of-distribution generalization, viewpoint robustness, and high interpretability, highlighting the potential of this unified goal-driven reasoning paradigm.


\section{Related Work}

\noindent \textbf{General Robot Manipulation Models.}
Developing a general robot manipulation policy has long been a topic of widespread
interest in the community.
The dominant paradigm, named VLA approach, employs end-to-end imitation learning that directly maps input observations and instructions to executable low-level actions~\cite{brohan2022rt, brohan2023rt, walke2023bridgedata, ebert2021bridge, huang2023embodied, li2023vision, zhen20243d, driess2023palm, chen2024moto,kim24openvla}.
The studies in this line can be divided into three main categories:
(i) Training from scratch.
For example, Octo~\cite{team2024octo} proposed a transformer-based action prediction pre-training and fine-tuning pipeline; Diffusion Policy~\cite{chi2023diffusion} explored the effectiveness of diffusion modeling in action prediction for undermining diverse and robust policies.
(ii) Building on video generation models or world models to transfer the abundant motion knowledge from real world into embodied policy prediction~\cite{chen2024moto, ye2024latent, chen2024igor, cui2024dynamo}.
(iii) Building on LMMs to extend the exceptional visual and language understanding capabilities of LMMs to action prediction.
For instance, OpenVLA~\cite{kim24openvla} trained an end-to-end VLA model by treating robot actions as tokens in the language model vocabulary; $\pi$0~\cite{black2024pi_0} introduced an additional diffusion module to directly predict normed continuous actions.

Another line of research proposed utilize LMMs to generate task-specific outputs, such as textual rules or constraint points, that can then be integrated with conventional motion solver~\cite{huang2023voxposer, wang2024dart, huang2024rekep, tang2025geomanip}.
Although these constraints are easy to interpret, the low-level solver struggles to leverage them due to the gap between high-level commands and specific robotic actions.

\noindent \textbf{Reasoning in Large Multimodal Models.}
The evolution of reasoning models has progressed from Chain-of-Thought (CoT) prompting~\cite{wei2022chain} to Reinforcement Learning (RL)-based self-motivation. 
OpenAI’s o1/o3 models~\cite{jaech2024openai} have sparked interest in leveraging long-form reasoning to enhance models’ ability to tackle complex tasks.
Recently, leaded by Deekseek-R1~\cite{guo2025deepseek}, researchers pay more attention to use reinforcement learning algorithms~\cite{watkins1992q,schulman2017proximal}, e.g. GPRO~\cite{shao2024deepseekmath}, to incentivize the model's reasoning abilities by leveraging the responses generated by model itself~\cite{yu2025dapoopensourcellmreinforcement,hu2025reinforce++,yang2025deepcritic}.

In the robot manipulation domain, several methods have also emerged to improve the accuracy of action predictions through reasoning mechanisms.
For instance, ECoT~\cite{zawalski2024robotic} introduces a templated reasoning process, prompting the model to analyze the current environment and produce a brief action plan before making predictions. 
HiRobot~\cite{shi2025hi} employs a dual-layer architecture that decomposes high-level user instructions into simpler, more direct commands for VLA execution.
However, these methods remain confined to System-1 level reasoning~\cite{kahneman2011thinking}, which is characterized by rapid and intuitive judgments based on patterns, thus failing to fully unleash the potential of deep reasoning.

In this work, we focus on how to incentivize models’ System-2 level reasoning capabilities~\cite{kahneman2011thinking, guo2025deepseek}, which are defined by deliberate and conscious thought processes with key features of self-reflection and verification behaviors, for realizing robot manipulation via step-by-step next-goal prediction.


\begin{figure}
    \centering
    \includegraphics[width=0.94\linewidth]{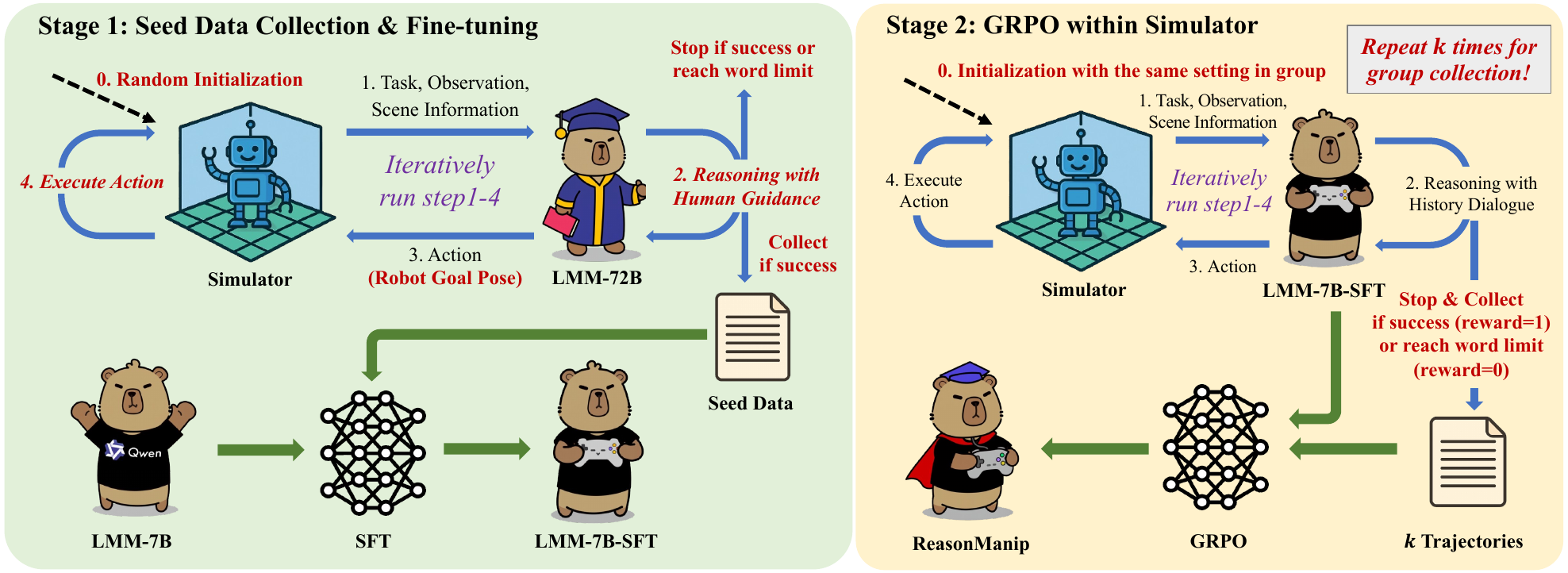}
    \vspace{-2pt}
    \caption{Overall pipeline of our method. On the first stage, we harness the advanced LMM-72B to generate robot manipulation reasoning data in the form of iteratively and interactive multi-round conversations with the users, which is used for SFT on a smaller LMM-7B model. On the second stage, the model interact with the virtual environment for GRPO training to further incentivize its reasoning ability for robot manipulation.}
    \label{fig:framework}
    \vspace{-15pt}
\end{figure}

\section{Method}
\subsection{Pipeline overview of \methodname}
In this work, we aim to leverage the exceptional reasoning capabilities of LMMs for realizing robot manipulation based on a novel task formulation.
Different from previous works that predict the low-level action offsets, we deploy the LMM to sequentially predict the next-goal gripper end pose, as shown in Fig.~\ref{fig:intro_paradigm}.
Based on this workflow, we first design a novel spatial reasoning formulation to help the LMM better understand the spatial action space.
Next, we propose a novel two-stage training framework to develop our \textbf{\methodname}, as illustrated in Fig.~\ref{fig:framework}.
In the first stage, we utilize advanced LMM to generate high-quality multi-round reasoning data by iteratively interacting with the environment under the guidance of brief human instructions for SFT.
Then, we perform GRPO within the simulator to further enhance the SFT model's reasoning ability for robotic manipulation.


In the following, we present our method into four parts: (i) spatial reasoning formulation in Subsec.~\ref{subsec:problem_reformulation}, (ii) user-guided multi-round reasoning data collection pipeline in Subsec.~\ref{subsec:interactive_collection}, (iii) the two-stage training framework in Subsec.~\ref{subsection:training}, and (iv) the scene information extraction strategy based on query-support matching for real-world deployment in Subsec.~\ref{subsection:segmentation}.
\subsection{Spatial Reasoning Formulation for Robotic Manipulation}
\label{subsec:problem_reformulation}

\begin{wrapfigure}[17]{r}{0.55\textwidth}
        \centering
        \vspace{-10pt}
        \includegraphics[width=0.99\linewidth]{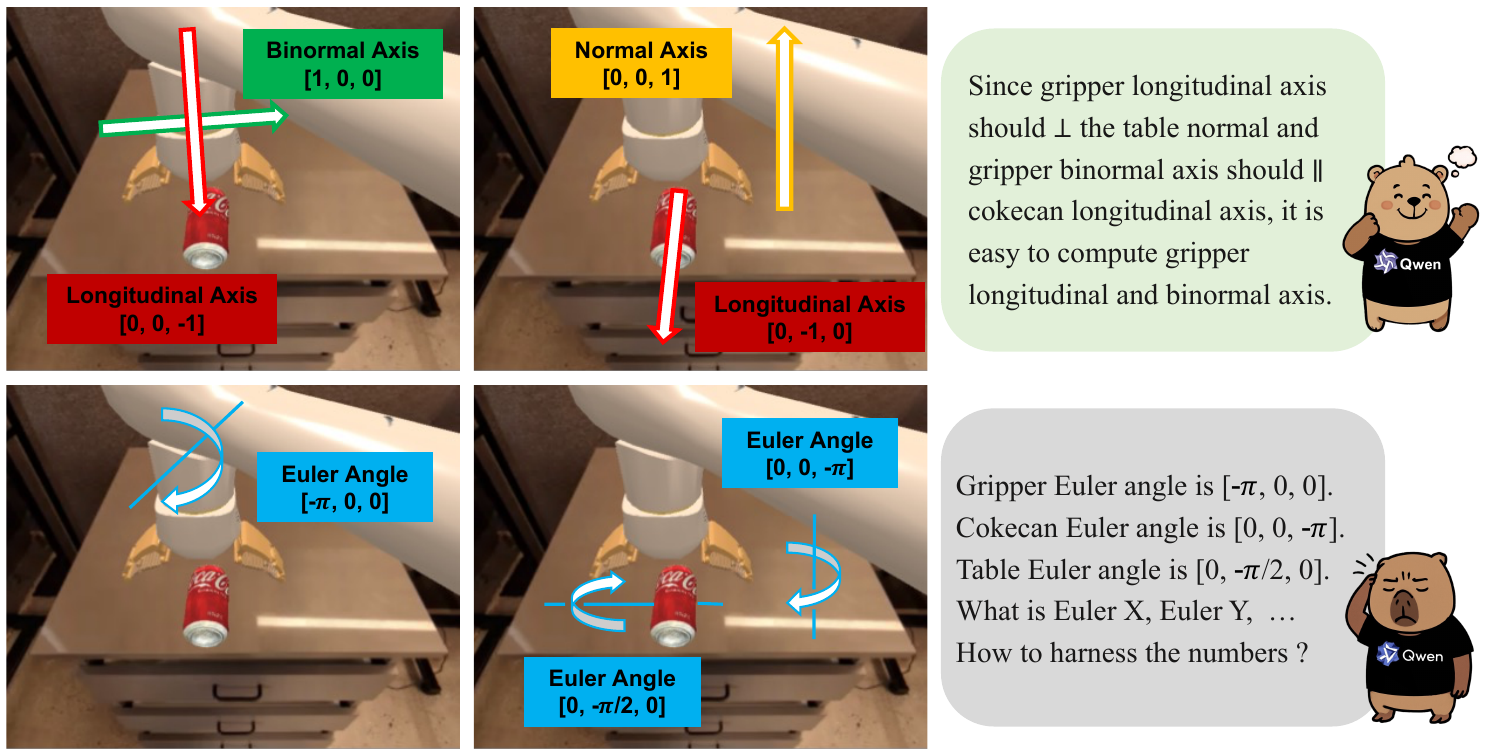}
        \caption{Example of the axis representation for the orientation of gripper (top left) and objects (top middle), which is in most cases more straightforward and manipulable for LMMs than the Euler angle representation (bottom left two subplots).}
        \label{fig:axis_representation}
\end{wrapfigure}

\textbf{Robot Manipulation in Multi-round Conversation Form.}\quad We employ the LMM to predict next-goal-wise gripper end pose in multi-round dialogue form for robotic manipulation.
Formally, for the $t^{\text{th}}$ round conversation, the LMM policy $\pi_\theta$ is expected to generate a response $\textbf{a}_t$ which consists of detailed reasoning of next-goal gripper end pose. 
The input contains the history conversations $\textbf{c}_{<t}=\{(\textbf{p}, \textbf{o}_{<t}, \textbf{s}_{<t}, \textbf{a}_{<t})\}$, current observation $\textbf{o}_{t}$, and current quantitative scene information $\textbf{s}_t$. $\textbf{o}_{<t}$ and $\textbf{s}_{<t}$ are the observations and scene information in the previous $t$ round conversations, respectively. 
$\textbf{p}$ is the basic query prompt that points out the task and response format (refer to Appx.~\ref{appendix:base_prompt}).
Then, the specific robot gripper pose solution $\textbf{sol}_t$ is extracted by parsing the response $\textbf{a}_t$. 
The gripper in environment is operated to reach this pose.
Next, the $t^{\text{th}}$ round conversation $(\textbf{o}_t, \textbf{s}_t, \textbf{a}_t)$ is appended to the history conversations for next-round generation. 
Overall, the $t^{th}$-round conversation is rigorously formulated as $\pi_\theta(\mathbf{a}_t|\mathbf{p}, \mathbf{o}_{\le t},\mathbf{s}_{\le t}, \mathbf{a}_{< t})$.
In this way, the process of completing a manipulation task is divided into sequential next-goal predictions and formulated in a multi-turn conversation form. We provide more examples illustrating the multi-turn conversations in Appx.~\ref{appendix: examples_multi_round_conversations}.

\textbf{Axis-based Spatial Representation.}\quad 
In this part, we introduce how to represent 
coordinates and axis-based rotation orientation of scene information $\textbf{s}_t$ and gripper pose $\textbf{sol}_t$.

First, $\textbf{s}_t=\{( (x, y , z), (l, w, h), (x_a, y_a, z_a), (x_b, y_b, z_b), (x_n, y_n, z_n))_i\}_{i=1}^N$, where $N$ is the number of object parts in the scenes. 
For each object part, this 12-dimensional tuple records its quantitative states. 
Specifically, $(x,y,z)$ denotes the object part's center position in Cartesian coordinates, while $(l,w,h)$ represents the bounding box dimensions, corresponding to (length, width, height). 
Notably, the remaining 9 elements are the \textbf{Axis Rotation Representation} that encodes orientation through three orthogonal unit vectors. 
The longitudinal axis $(x_a, y_b, z_a)$ indicates the primary object extent, which is obtained by taking the eigenvector with largest eigenvalues of the object part point cloud PCA result. 
The normal axis $(x_n, y_n, z_n)$ is orthogonal to dominant surfaces, which is obtained by taking the eigenvector the the smallest eigenvalues of the object point cloud PCA result. 
And the binormal axis is calculated by $(x_b, y_b, z_b) = (x_a, y_a, z_a)\times(x_n, y_n, z_n)$. 
Examples of the axis representation (top subplots) and the Euler angles (bottom subplots) are illustrated in Fig.~\ref{fig:axis_representation}. 

Next, after obtaining the LMM response $\textbf{a}_t$ at $t^{th}$ round, the valid executable robot pose $\textbf{sol}_t=((x, y, z), (x^g_a, y^g_a, z^g_a), (x^g_b, y^g_b, z^g_b), gripper)$ are parsed from it. 
($x,y,z$) is the coordinate of the gripper position. ($x^g_a, y^g_a, z^g_a$) is the longitudinal axis of the gripper, which also represents the gripper pointing direction. ($x^g_b, y^g_b, z^g_b$) is the binormal vector of the gripper, which also represents the gripper release/grasp direction. $gripper$ is a binary value indicating the open/close state of the gripper. 
When performing actions in environment, the gripper rotation matrix $\mathbf{R}$ is calculated by
\begin{equation}
    \mathbf{R} = 
\begin{bmatrix} x_a \\ y_a \\ z_a \end{bmatrix}
\begin{bmatrix} x^0_{a} & y^0_{a} & z^0_{a} \end{bmatrix}
+
\begin{bmatrix} x_b \\ y_b \\ z_b \end{bmatrix}
\begin{bmatrix} x^0_{b} & y^0_{b} & z^0_{b} \end{bmatrix}
+
\left( \begin{bmatrix} x_a \\ y_a \\ z_a \end{bmatrix} \times \begin{bmatrix} x_b \\ y_b \\ z_b \end{bmatrix} \right)
\left( \begin{bmatrix} x^0_{a} \\ y^0_{a} \\ z^0_{a} \end{bmatrix} \times \begin{bmatrix} x^0_{b} \\ y^0_{b} \\ z^0_{b} \end{bmatrix} \right)^\top
\end{equation}
where $(x_{a}^{0}, y_{a}^{0}, z_{a}^{0})$ and $(x_{b}^{0}, y_{b}^{0}, z_{b}^{0})$ are gripper initial longitudinal axis and binormal axis, respectively. 

\subsection{User-guided Multi-Round Reasoning Data Collection}
\label{subsec:interactive_collection}
\begin{figure}[t]
    \centering
    \includegraphics[width=\linewidth]{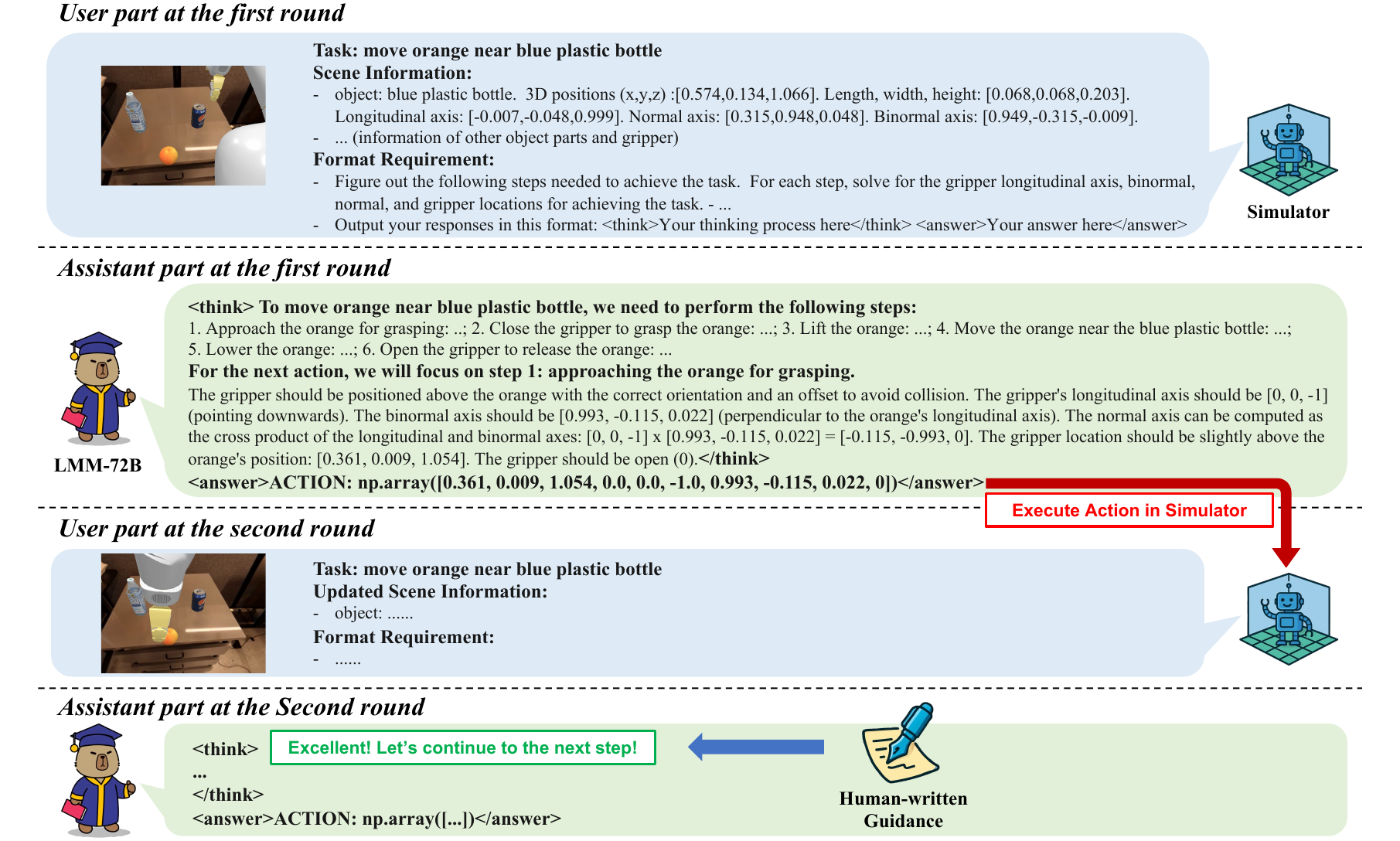}
    \caption{The illustration of our multi-round conversation collection pipeline. The predicted gripper pose is executed in a simulator to obtain the updated observation and scene information. We utilize human-written guidance to help model reason and generate the next-goal prediction.}
    \label{fig:data-collection}
    \vspace{-10pt}
\end{figure}

Fig~\ref{fig:data-collection} illustrates the pipeline we designed for multi-round reasoning data collection.
At each turn $t$, the basic prompt $\textbf{p}$, the current observation $\textbf{o}_t$ and the quantitative scene information $\textbf{s}_t$ are collected from the simulator.
We first ask the user to write a simple instruction $\textbf{u}_t$ based on the previous assistant's response. This instruction may provide feedback on the assistant's answer (e.g.,  "We are doing good, continue"), directly instruct the next move (e.g., "Move over to the orange"), or reflect based on the scene observation (e.g., "The cokecan is knocked over because the gripper is too low").
We then ask the existing Qwen2.5-VL-72B-Instruct~\cite{Qwen2.5-VL} to reason and compute the next-goal gripper end pose as $\textbf{a}_t$. 
Formally, $\textbf{a}_{t}=LMM(\textbf{p}, \textbf{o}_{\le t},\textbf{s}_{\le t},\textbf{a}_{<t}, \textbf{u}_t)$.
Next, we extract robot pose solution $\textbf{sol}_t$ from the assistant response $\textbf{a}_t$ and move the gripper in the simulator.
The updated quantitative scene information $\textbf{s}_{t+1}$ and the updated visual observation $\textbf{o}_{t+1}$ are then captured for the next-turn conversation.
We only collect the conversations that successfully command the robot to complete the task. After completion, we revert the role in the conversation for $\textbf{u}_t$ from ``user'' to ``assistant'' and integrate it with the next step reasoning $\textbf{a}_t$, denoted as $\textbf{a}_{t+1}\leftarrow [\textbf{u}_t, \textbf{a}_{t+1}]$. In this way, the user's instruction becomes part of the assistant reasoning process, which incentivizes the LLM to reason in a "self-talk" and "self-reflection" manner.

\subsection{Teaching LMM to reason for robot manipulation}
\label{subsection:training}

\textbf{Supervised Fine-tuning.}
First, we denote history conversations $\mathbf{c}_{\le t}=(\mathbf{p},\mathbf{o}_{\le t},\mathbf{s}_{\le t},\mathbf{a}_{< t})$, which represents for the basic prompt $\textbf{p}$, history conversations $(\textbf{s}_{<t},\textbf{o}_{<t},\textbf{a}_{<t})$, current scene information $\mathbf{s}_t$, and current observation $\mathbf{o}_t$. Given a dataset $\mathcal{D} = \{(\mathbf{c}_{\le t}, \mathbf{a}_{ t})_{n}\}_{t=1,n=1}^{T,N}$, we train a policy $\pi_\theta$ by minimizing the cross entropy loss:
\begin{equation}
\mathcal{L}_{\text{SFT}}(\theta) = \frac{1}{T}\sum_{t=0}^T\mathbb{E}_{(\mathbf{c}_{\le t}, \mathbf{a}_{t}) \sim \mathcal{D}} \frac{1}{L}\sum_{i=0}^L\left[ -\textbf{log}p(\mathbf{a}_{t, i}|\pi_\theta(\mathbf{c}_{\le t}, \mathbf{a}_{t, < i})) \right],
\end{equation}
where $\theta$ denotes the policy parameters, $t$ denotes the $t^{\text{th}}$ turn of the $T$-turn  conversations, $i$ is the token index of $a_t$ consists of $L$ tokens in total.

\textbf{Group Relative Policy Optimization within Simulator.}
During GRPO training, given the specific task, the model $\pi_\theta$ interacts with the environment $K$ times with the same initial settings to generate $K$ $T$-round conversations $\{(\mathbf{p}, \textbf{o}_{\le t}, \textbf{s}_{\le t}, \textbf{a}_{\le t})_{k}\}_{k=1}^K$. We extract answer $\textbf{sol}_{t}$ from $\textbf{a}_t$ and the robot executes it. Consequently, the simulator evaluates after the $t^\text{th}$ conservation of each of the $K$ trajectory and generates rewards $r_k=\mathbbm{1}(\text{The task completes}|\textbf{sol}_1,...,\textbf{sol}_t)$, which is a binary number indicating the success (1) or failure (0) of the task. The GRPO objective maximizes:
\begin{equation}
\mathcal{L}_{\text{GRPO}}(\theta) =  \frac{1}{T}\sum_{t=0}^T\mathbb{E}_{\mathbf{a}_{t} \sim \pi_{old}(O|\mathbf{c}_{\le t})} \left[\frac{1}{L}\sum_{i=1}^L\frac{1}{K}\sum_{k=1}^K \min\left( \frac{\pi_\theta(\mathbf{a}_{t.i}|\mathbf{c}_{\le t}, \mathbf{a}_{t,< i})}{\pi_{old}(\mathbf{a}_{t.i}|\mathbf{c}_{\le t}, \mathbf{a}_{t,< i})} \hat{A}_k, clip(\epsilon, \hat{A}_k) \right) \right],
\end{equation}
where $clip(\epsilon, \hat{A}) = \text{clip}\left( \frac{\pi_\theta}{\pi_{\text{old}}}, 1-\epsilon, 1+\epsilon \right) \hat{A}$ is the clipping operation; $\hat{A}_k$ denotes the advantage value of the $k^{\text{th}}$ sampled output, which is calculated as $\hat{A}_k = \frac{r_k-\text{mean}(r_k)}{\text{std}(r_k)}$; 
$\text{mean}(\cdot)$ and $\text{std}(\cdot)$ are the mean and standard deviation of the $K$ rewards, respectively; $\pi_{old}$ is the reference policy, $K$ is the number of explored sample;
and $clip(\cdot)$ confines the change of the new policy $\pi$ relative to the reference policy with parameter $\epsilon$.

\vspace{-5pt}
\subsection{Quantitative Scene Information Extraction}
\vspace{-5pt}
\label{subsection:segmentation}
To obtain the quantitative scene information $s_t$ at $t^{th}$ round, it is necessary to segment each object part $i$ and determine its current position, scale, and axis representation.
\methodname first predicts the object parts involved for the next-goal prediction.
Next, for each object part, we use an open-world part segmentation network to generate the segmentation mask. 
The segmentation mask is then projected onto the point cloud to calculate the above states. 
However, although there are some open-world part segmentation works~\cite{liang2023open, lai2024lisa, fineclip}, their performance in robot manipulation scenarios remains unsatisfactory~\cite{tang2025geomanip}.
%
%

To address this challenge, we use a support query segmentation matching method and a database query to achieve open-world part segmentation in real-time.
Specifically, we first build a large database $\mathcal{DB}:part \rightarrow \{(I^s,M^s)_i\}_i^N$, where $part$ is the description of the object part to be segmented, $I^s$ is one example image containing the corresponding object part (support image), $M^s$ is corresponding segmentation mask (support mask), and $\{(I^s,M^s)^N_i\}_i$ is $N$ support image - segmentation mask pair. Next, we predict the segmentation mask $M^q$ of the object part description $part$ in the query image $I^q$ by a model $M^q=Segm(I^q;\{(I^s,M^s)_i\}_i^N\})$. Generally, the model $Segm$ use a shared backbone to extract the support features and query features. Support features are refined with attention design with the query feature before mask prediction. 
Please refer to Appx.~\ref{appendix:object_part_segm} for network details.
\begin{figure}[t]
    \centering
    \begin{minipage}[t]{0.76\textwidth}
        \centering
        \includegraphics[width=\linewidth]{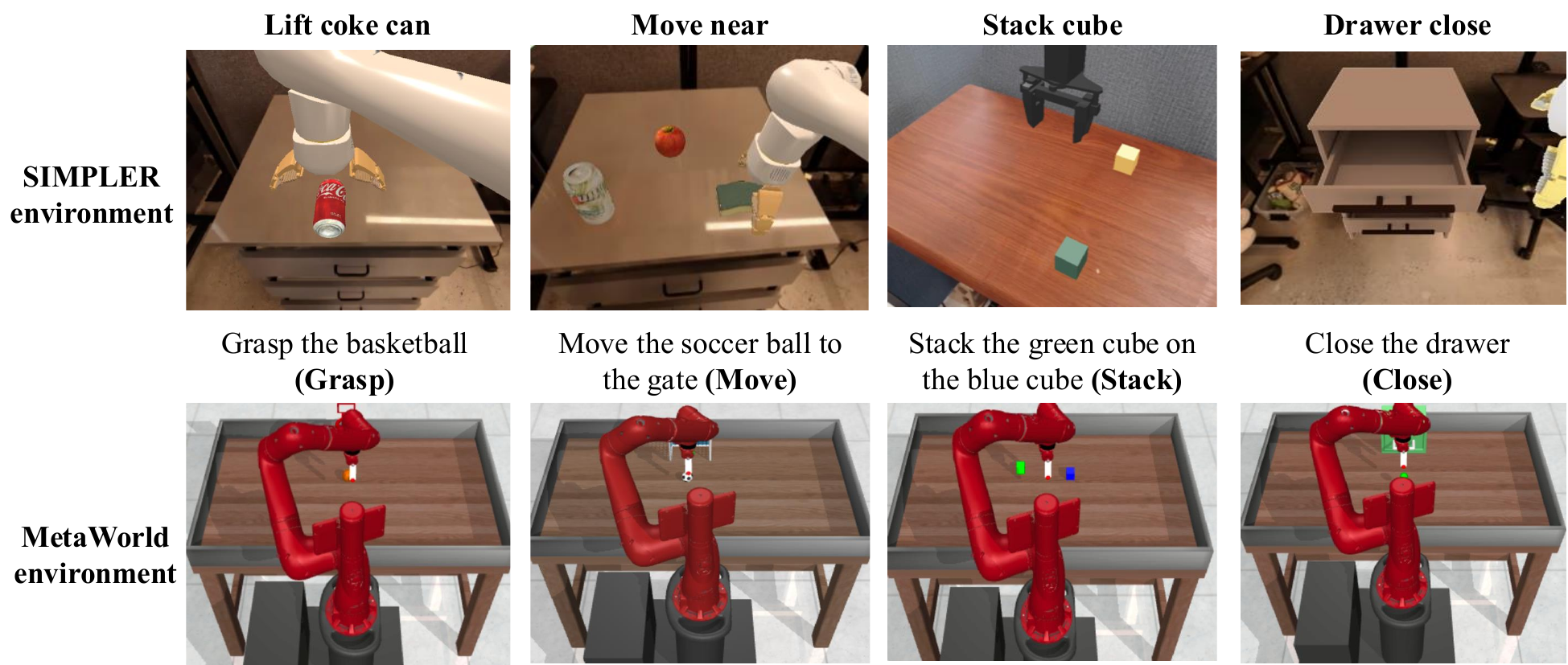}
        \caption{Illustration of the environment comparisons of the SIMPLER and the MetaWorld environment. We can see that they have large gaps in camera views, robots, object visuals, and backgrounds.}
        \label{fig:env_gap}
    \end{minipage}
    \hfill
    \begin{minipage}[t]{0.19\textwidth}
        \centering
        \includegraphics[width=\linewidth]{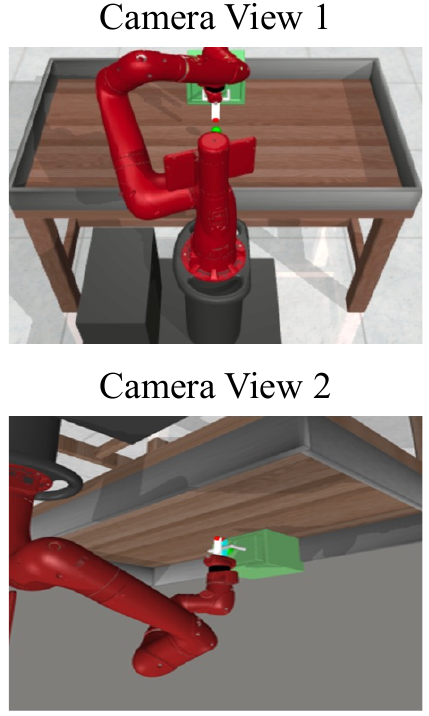}
        \caption{Different camera views in MetaWorld.}
        \label{fig:camera_view}
    \end{minipage}
    \vspace{-12pt} 
\end{figure}
\section{Experiments}

%
%
%

\subsection{Implementation Details}
We use the Qwen2.5-VL-7B-Instruct~\cite{yang2024qwen2} as our base LMM model.
We collect 5-10 samples per task in the SIMPLER environment, resulting in lightweight 65 training samples in total across 7 tasks as the seed data.
The training details of SFT and GRPO are included in the Appx.~\ref{appendix:exp_detail}.

%

%
%
\subsection{Experiments on Virtual Environment}
Our method is compared against four existing approaches: (i) \textbf{RT-1-X}~\cite{o2023open,brohan2022rt}: a large-scale vision-language-action model with 55B parameters in a trajectory transformer architecture;
(ii) \textbf{Octo-base}~\cite{team2024octo}: an open-source generalist policy with 93M parameters trained on diverse robot demonstrations.
(iii) \textbf{OpenVLA}~\cite{kim24openvla}: a 7B-parameter open-source VLA model combining based on Llama-2~\cite{touvron2023llama}.
(iv) \textbf{Moto}~\cite{chen2024moto}: a modular architecture with hidden action tokens learned with videos.
In addition, we evaluate two variants of our method, (i) \textbf{SFT}, the Supervised Fine-Tuning version trained on task-specific demonstrations, and (ii) \textbf{SFT+GRPO}, which involves further training the model with GRPO, serving as our full model.

We evaluate our method across seven tasks in the SIMPLER environment~\cite{li24simpler}. The average success rates of 50 trials for each task are presented in Table \ref{tab:benchmark_results}, which indicates that \methodname exhibits competitive performance with existing methods in most scenarios. 
Our SFT model demonstrates exceptional outcomes, achieving particularly high success rates in tasks such as Lift coke can (77\%) and Put carrot on plate (34\%). 
Notably, in contrast to VLA policies that typically require millions of training demonstrations, \methodname achieves strong performance with minimal data, highlighting its remarkable data efficiency.
In addition, the introduction of GRPO further enhances its performance, achieving particularly high success rates in tasks such as Move near (67\%) and Stack cube (57\%).

\begin{table}[t]
\centering
\caption{Performance comparison on SIMPLER benchmark tasks (success rate \%).}
\label{tab:benchmark_results}
\resizebox{0.99\textwidth}{!}{
\small
\setlength{\tabcolsep}{4.5pt}
\begin{tabular}{l|ccccccc|c}
\toprule
 & \textbf{Lift coke can} & \textbf{Move near} & \textbf{Stack cube} & \begin{tabular}{@{}c@{}}\textbf{Put carrot} \\ \textbf{on plate}\end{tabular} & \begin{tabular}{@{}c@{}}\textbf{Put spoon} \\ \textbf{on towel}\end{tabular} & \textbf{Drawer open} & \textbf{Drawer close} & \textbf{Average} \\
\midrule
RT-1-X & 56.7 & 31.7 & 0 & 4.2 & 0 & \textbf{51.9} & \textbf{74.1} & 31.2 \\
Octo-base & 17.0 & 4.2 & 31.9 & 8.3 & 12.5 & 14.8 & 51.9 & 20.1 \\
OpenVLA & 16.3 & 46.2 & 0 & 0 & 0 & 15.8 & 19.5 & 14.0 \\
Moto & 74.0 & 60.4 & -- & -- & -- & 13.0 & 73.2 & - \\
\midrule
\textbf{Ours (SFT)} & \textbf{77.0} & 30.0 & 55.0 & \textbf{34.0} & 32.0 & 10.0 & 42.0 & 40.0 \\
\textbf{Ours (+GRPO)} & 72.0 & \textbf{67.0} & \textbf{57.0} & 30.0 & \textbf{40.0} &10.0  &42.0 & \textbf{45.4}\\
\bottomrule
\end{tabular}
\vspace{-25pt}
}
\end{table}

\begin{figure}
    \centering
    \includegraphics[width=\linewidth,height=0.34\linewidth]{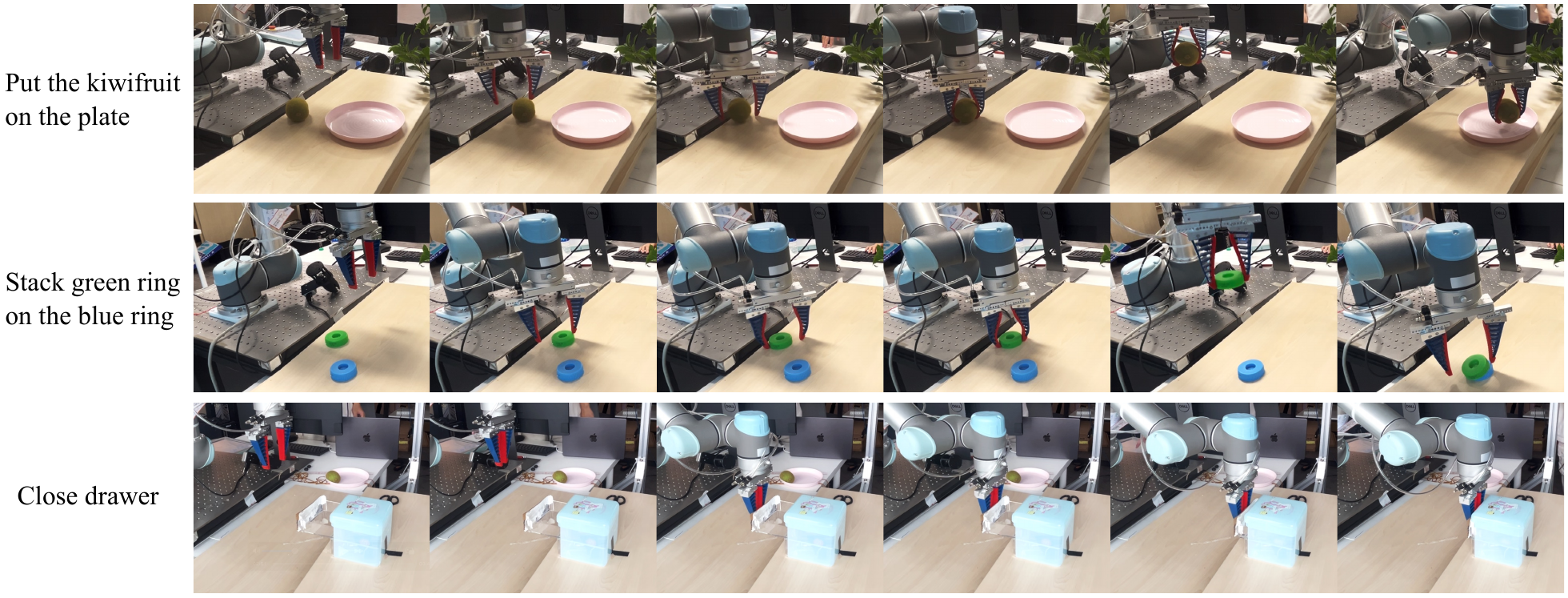}
    \caption{Sequence of our method's execution in the real-world environment for the three tasks.}
    \label{fig:seq_rw}
    \vspace{-10pt}
\end{figure}

\subsection{Zero-shot generalization to Significant Different Unseen Environment}
We demonstrate the strong generalizability of our method by deploying \methodname, trained on the SIMPLER simulator, directly to the unseen MetaWorld~\cite{yu2020meta} environment without fine-tuning.
We evaluate our method against existing approaches across four tasks resembling those in SIMPLER: Grasp the basketball (Grasp), Move the soccer ball to the gate (Move), Stack the green cube on the blue cube (Stack), and Close drawer (Close). Fig.~\ref{fig:env_gap} visually compares these similar tasks in SIMPLER and MetaWorld, highlighting their significant differences in camera views, robots, object shapes, and backgrounds.

\begin{table}[t]
\centering
\begin{minipage}{0.48\textwidth}
\setlength{\tabcolsep}{4pt}
\centering
\caption{Success Rates (\%) of Existing Methods and Ours on the OOD MetaWorld Benchmark.}
\resizebox{0.98\textwidth}{!}{
\begin{tabular}{l|cccc}
\toprule
 & \textbf{Grasp} & \textbf{Move} & \textbf{Stack} & \textbf{Close} \\
\midrule
RT-1-X & 0 & 0 & 0 & 0 \\
Octo-base & 0 & 0 & 0 & 0 \\
OpenVLA & 0 & 0 & 0 & 0 \\
\hline
\textbf{Ours (SFT)} & \textbf{100.0} & 23.3 & \textbf{20.0} & 20.0 \\
\textbf{Ours (+GRPO)} & 96.7 & \textbf{30.0} & \textbf{20.0} & \textbf{40.0} \\ 
\bottomrule
\end{tabular}
}
\label{tab:env_gap}
\end{minipage}
\hfill
\begin{minipage}{0.48\textwidth}
\centering
\caption{Success Rates (\%) of our method and existing approaches on real-world setups. $*$ denotes the model is fine-tuned on real-world scenes.}
\resizebox{0.9\textwidth}{!}{
\begin{tabular}{l|ccc}
\toprule
 & \textbf{Stack} & \textbf{Put On} & \textbf{Drawer} \\
\midrule
RT-1-X & 0 & 0 & 0  \\
OpenVLA* & 0 & 0 & 0  \\
$\pi 0$* & 0 & 0 & 20.0 \\
\hline
\textbf{Ours} & \textbf{40.0} & \textbf{50.0} & \textbf{50.0} \\
\bottomrule
\end{tabular}
}
\label{tab:realworld}
\end{minipage}
\vspace{-10pt}
\end{table}

We compare our method with three existing approaches: RT-1-X~\cite{o2023open,brohan2022rt}, Octo-base~\cite{team2024octo}, and OpenVLA~\cite{kim24openvla}. 
For each task, we conduct 30 trials and report the average success rates. Our method outperforms existing approaches in most SIMPLER scenes (Tab.~\ref{tab:benchmark_results}), and the results in Tab. 
Tab.~\ref{tab:env_gap} shows an even larger performance gap between our method and the existing approaches, demonstrating the superior generalization to unseen environments of our model.

What's more, we change the camera view in the MetaWorld environment significantly, as illustrated in Fig.~\ref{fig:camera_view}, and evaluate these four tasks again. 
According to the Tab.~\ref{tab:camera_view} we observe that the change of camera viewpoint makes extremely little impact on \methodname's performance, which shows that our method performs strong viewpoint robustness.

This strong out-of-distribution generalization ability of our method stems from two key characteristics: (i) the mathematical formulation of manipulation tasks inherently captures physical invariants that remain consistent across various domains, and (ii) our axis-based representation enables better spatial reasoning and avoids dependencies on visual appearance.

Compared to using supervised fine-tuning (SFT) alone, integrating GRPO results in a large performance gain in out-of-distribution (OOD) settings, as shown in Tab.~\ref{tab:env_gap}.
This improvement is partly because of the exploration-exploitation dynamics of GRPO, which actively mitigate the overfitting tendency of SFT  when trained on limited data.
Additionally, GRPO's reward learning prioritizes fundamental physical principles over specific manipulation trajectories, resulting in policies that emphasize core manipulation logic and foster more adaptable and generalizable strategies.
%

\subsection{Real World Experiment}
%
%
To demonstrate the effectiveness of our method in real-world scenarios, we deploy our trained model on a robotic platform equipped with a RealSense 435i camera for visual perception and a UR5 robot arm for execution. 
We assess the performance of \methodname in comparison with three state-of-the-art approaches: \textbf{RT-1-X}~\cite{o2023open,brohan2022rt}, \textbf{OpenVLA}~\cite{kim24openvla}, and \textbf{$\pi 0$}~\cite{black2024pi_0}, where the latter two models are fine-tuned on 50 manually collected ground-truth trajectory data. 
The experimental setup and the training details are provided in Appx.~\ref{appendix:exp_detail}.
We establish three tasks for evaluation: (i) Stack the green lip onto the blue lip (Stack), (ii) Put the kiwi onto the plate (Put On), and (iii) Close the drawer (Drawer). 
For each task, we conduct 10 experiments and report average success rates in Tab.~\ref{tab:realworld}.
Our results show that our method consistently outperforms the compared approaches across all three tasks, even when OpenVLA is fine-tuned with ground-truth data. 
Visual results of our execution are presented in Fig.~\ref{fig:seq_rw}.
These demonstrate our method's generalizability in the sim-to-real settings.

\begin{figure}[t]
    \centering
    \includegraphics[width=\linewidth]{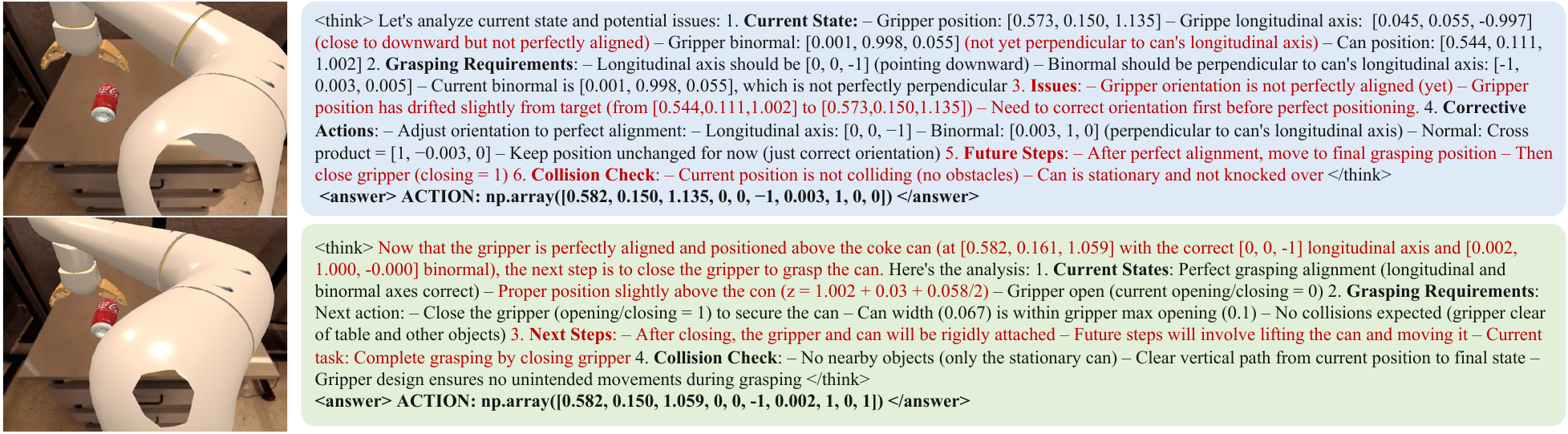}
    \caption{Example of conversation demonstrating the system-2 reasoning ability.}
    \label{fig:system-2}
    \vspace{-10pt}
\end{figure}

\begin{table}[t]
\centering
\begin{minipage}{0.48\textwidth}
\centering
\caption{Success Rates (\%) of Our Method on MetaWorld under Different Camera View}
\begin{tabular}{l|cccc}
\toprule
 & \textbf{Grasp} & \textbf{Move} & \textbf{Stack} & \textbf{Close} \\
\midrule
View 1 & 96.7 & 30.0 & 20.0 & 40.0 \\
View 2 & 100.0 & 26.7 & 20.0 & 50.0 \\ 
\bottomrule
\end{tabular}
\label{tab:camera_view}
\end{minipage}
\hfill
\begin{minipage}{0.48\textwidth}
\centering
\caption{Success Rates (\%) Comparisons using Euler Representation and Axis Representation}
\setlength{\tabcolsep}{1.5pt}
\resizebox{0.99\textwidth}{!}{
\begin{tabular}{l|cccc}
\toprule
 &\textbf{Lift cokecan} & \textbf{Move near} & \textbf{Drawer} \\
\midrule
Ours(Euler) & 42.0 & 28.0 & 40.0   \\
Ours(Axis)& 96.7 & 55.0 & 42.0   \\
\bottomrule
\end{tabular}}
\label{tab:axis_vs_euler}
\end{minipage}
\vspace{-10pt}
\end{table}

\subsection{Discussion and Analysis}
\vspace{-5pt}
\textbf{How does the system-2 reasoning process help in successful manipulation?}
We discover that with the system-2 reasoning process, the LMM obtains two typical features: (i) Failure detection; (ii) Future condition discussion and reasoning. 
See example in Fig.~\ref{fig:system-2}, \methodname detects possible misalignment and corrects it for a safer grasp, and then discusses the future steps. These reasoning behaviors ensure a safer and more robust manipulation process, which also directly contributes on realizing accurate goal pose estimation.
Please refer to the Appx.~\ref{appendix: examples_multi_round_conversations} for more examples.

\textbf{How does our method benefit from the axis rotation representation?} In order to prove that the LMM model is better to understand and harness the axis representation for spatial reasoning compared to the Euler representation, we reformat all the orientations in the training data to the Euler representation. We then apply SFT with the same settings, and test its performance in the same environments for the three tasks (Lift coke can, Move near, and Drawer close). The results are shown in Tab.~\ref{tab:axis_vs_euler}, and we can see that the performance drops significantly using the Euler representation. We provide inference comparisons using the Euler angle and axis representation in the Appx.~\ref{appendix:euler_vs_axis}

\textbf{How is the efficiency of the reasoning methods compared to end-to-end VLA models?} A common trade-off in reasoning-based models is the increased computational cost required to spend in reasoning and achieve higher accuracy. In contrast, traditional Vision-Language-Action (VLA) models are widely adopted for their simple architecture and high-frequency control capabilities. To demonstrate the time efficiency of our approach, we record the overall time spent on completing the task \textit{Put On}, and it takes only around 40 seconds on average, which maintains competitive inference speeds despite incorporating the reasoning paradigm. 

\section{Conclusion}

In this work, we propose \methodname, a novel robot manipulation paradigm of direct next-goal pose prediction via unified system-2 reasoning of LMM. To make the LMM better understand the spatial action space, we design a novel task formulation and axis rotation representation. Next, we introduce a user-guided data collection pipeline and an effective two-stage training framework to incentivize the LMM models' reasoning ability for robot manipulation. Extensive experiments on both virtual and real-world environments demonstrate the data efficiency, out-of-distribution generalizability, viewpoint robustness, and interpretability of our ReasonManip.

\newpage
\bibliographystyle{plainnat}
\bibliography{reference.bib}

\newpage
\appendix
\section{Implementation Details of the Experiments}
\label{appendix:exp_detail}

\textbf{Training Details}
We use the Qwen2.5-VL-7B-Instruct~\cite{yang2024qwen2} as our base LMM model.
We collect only 5-10 samples per task in the SIMPLER environment, resulting in 65 training samples in total across 7 tasks used for SFT.
During SFT, we specify the learning rate as $1 \times 10^{-5}$ and use the AdamW~\cite{loshchilov2017decoupled} optimizer with global batchsize of $32$. We train it for 15 epochs by using 8 A800 GPUs.
For GRPO training, update iteration is set to 1; the learning rate is set to $5 \times 10^{-6}$; and the cosine annealing scheduler~\cite{loshchilov2016sgdr} is used. 
We use the GRPO to train for each of the specific tasks, since we observe severe task conflict based on the limited quantity of SFT training samples. Specially, for the three tasks "put on" task, "Put spoon on towel", "Put carrot on plate", and "Stack cube", which share similar manipulation logic, we use the GRPO training to train the mixed scenes together.

For fine-tuning the comparable methods OpenVLA and $\pi 0$ for real-world applications, we collect 50 ground-truth trajectories and use LoRA~\cite{hu2022lora} to fine-tune them for 20 epochs. For OpenVLA, the LoRA rank is set to 64, the batch size is set to 8, and the learning rate is set to 5e-4. For $\pi 0$, the batch size is set to 32, the action horizon is set to 10, and the learning rate is set to 1e-5.

\textbf{Details settings of the virtual environment}
For the SIMPLER environment, since there are unreachable gripper positions in the original object placement settings, we adjust the object position in the xy-plane slightly to ensure the gripper's reachability. Specifically, for the task "Lift coke can" and "Move near", the placement range of the objects are shifted by -0.1m in the x coordinate. For the task "Drawer open" and "Drawer close", the placement range of the objects is shifted by 0.1m in the x coordinate. For the MetaWorld environment, we use the default camera view setting. We set camera\_idx as 1 for the different views.

\begin{figure}[h]
    \centering
    \includegraphics[width=0.8\linewidth]{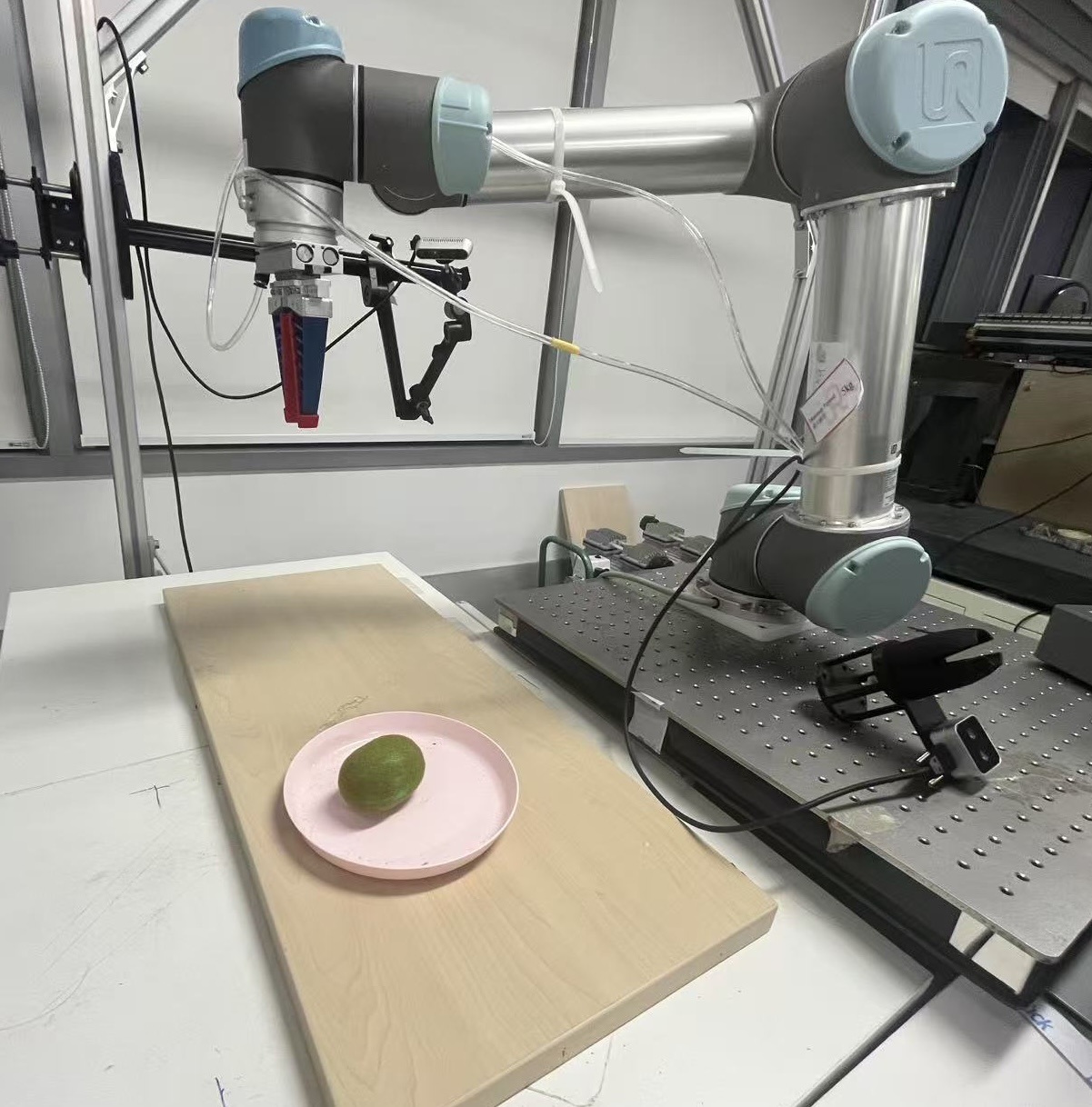}
    \caption{Illustration of the real-world experiment settings.}
    \label{fig:rw_setting}
\end{figure}

\textbf{Details settings of the real-world environment}
We use an external Realsense 435i camera to capture the RGBD information of the scene. We use the UR5 robot for manipulation. We use a single A100 GPU for model inference. The real-world experiment setting can be viewed in Fig.~\ref{fig:rw_setting}

\section{Details of Support-Query Matching-Based Object Part Segmentation}
\label{appendix:object_part_segm}
The support-query matching-based object part segmentation generate an object part mask given the object part description. It consists of two stages. On the first stage, we query a database given the object part description as the query to retrieve the corresponding support images and the support masks for the object part. On the second stage, a support-query matching-based segmentation model takes the support images, support masks and the query image as the input, and generate mask prediction for the query image. 

\textbf{Database Design}\quad The database $\mathcal{DB}:part \rightarrow \{(I^s,M^s)_i\}_i^N$, where $part$ is the description of the object part to be segmented, $I^s$ is one example image containing the corresponding object part (support image), $M^s$ is corresponding segmentation mask (support mask), and $\{(I^s,M^s)^N_i\}_i$ is $N$ support image - segmentation mask pair. Although there are many better methods for retrieval, we use a very naive approach for simple implementation. We use the description of the object part in the database as the key, $\{(I^s,M^s)^N_i\}_i$ as the value, and $part$ as the query. We retrieve $\{(I^s,M^s)^N_i\}_i$ whose key (object part description) hits the query ($part$) the most. 

\textbf{Support-query matching-based segmentation network}\quad Next, we predict the segmentation mask $M^q$ of the object part description $part$ in the query image $I^q$ by a model $M^q=Segm(I^q;\{(I^s,M^s)_i\}_i^N\})$. Specifically, the model $Segm$ first use a shared backbone (Res-50~\cite{he2016deep}, Res-101~\cite{he2016deep}, or Swin-B~\cite{liu2021swin}) to extract all the intermediate support features $X^1_s,X^2_s,\cdots X^L_s$ and intermediate query features $x^1_q,x^2_q,\cdots x_q^L$, where $X^l_s\in\mathbb{R}^{N_s\times H_l\times W_l\times D_l}$ and $q^l_s\in \mathbb{R}^{H_l\times W_l\times D_l}$. $H_l,W_l,D_l$ are the $l^{\text{th}}$ layer feature map's height, width, and feature dimension, respectively. $N_S$ is the number of support features for the support set $S$. We group these features by aggregating the neighbour layers of features. Without the abuse of symbols, we still denote the support features and query feature after grouping as $X^1_s,\cdots X_s^{L}$ and $x_q^1,\cdots x_q^L$, respectively. For each layer $l$, the support mask $M^l_s$ is obtained by rescaling the original support mask to the shape $[H_l, W_l]$. Next, we adopt the Support Image Pruning~\cite{tang2025overcoming} to retrieve a small subset $S'$ from $S$. 
Specifically, in this step, the retrieval principle is
\begin{equation}
\begin{aligned}
        S' &= \mathop{\arg \max}_{{S'}\subset{S}} 
    \frac{1}{L}\sum_{l=1}^L \theta(X^l_{S'}), |S'|=N'.
\end{aligned}
\end{equation}
Where $X^l_{s'}$ are the $l^{\text{th}}$ layer of features of the retrieved support features and $\theta(X_{S'})$ measures the "contribution" of the support features to the query feature defined as:
\begin{equation}
\begin{aligned}
\label{Pruning2st}
    \theta(X_{S'}) =
    \sum_{i=1}^{N'}
    f(\frac{1}{|x_{{S'}_i}|}\sum_{j=1}^{|x_{{S'}_i}|}x_{{{S'}_i}_{[j]}})f(\frac{1}{|x_q|}\sum_{j=1}^{|x_q|}x_{q_{[j]}}).
\end{aligned}
\end{equation}
The definition of $f$ can refer to the Eq.~\ref{SymAtt3}. Next, we use the symmetric correlation modules metioned in~\cite{tang2025overcoming} to calculate the correlation scores between support features $X_{S'}$ and query feature $x_q$:
\begin{equation}
\begin{aligned}
\label{SymAtt3}
 A_i^l(X_{S'}^l,x_q^l)=\left[\text{softmax}(\frac{f^l(x_q^l)f^l(X_{S'}^l)^T}{d}))\right]_{(:,[\text{head}_i:\text{tail}_i])}, &\\
    f^l(x) = \frac{f_1^l(x)f_2^l(x)}{\|f_2^l(x)\|_2},i=1,...,N', &
\end{aligned}
\end{equation}
The coarse predicted mask at level $l$ is generated by  $C^l = A^l \cdot M^l$.
Next, we design a refiner that harnesses the top-down fusion to aggregate coarse masks of neighbour layers.
In each top-down step, we apply bilinear interpolation $U^{l-1} = \text{Upsample}(C^l)$ to align $C^l$ with the size of $C^{l-1}$, then we refine $C^{l-1}$ by a 2D convolution $F^{l-1} = \text{conv}^{l-1}(\text{concat}[U^{l-1},C^{l-1}])$.

We repeat the top-down process to fuse two consecutive layers' coarse predictions until we obtain the second-last-layer $F^{2}$. 
In the last step, we obtain the final binary output $F^1$ by $F^1=\text{conv}^1(\text{concat}[\text{Upsample}(F^2), x_q^1, \text{AvgPool}(X_{S'}^1)])$.

\section{Examples of Euler Angle Representations Failure Cases}
\label{appendix:euler_vs_axis}
In this section, we illustrate some failure cases using the Euler angle representations. From these examples, we can see that LMM fails to achieve the task since it can't harness its spatial reasoning ability to generate correct gripper orientation. 

\begin{figure}[h]
    \centering
    \includegraphics[width=\linewidth]{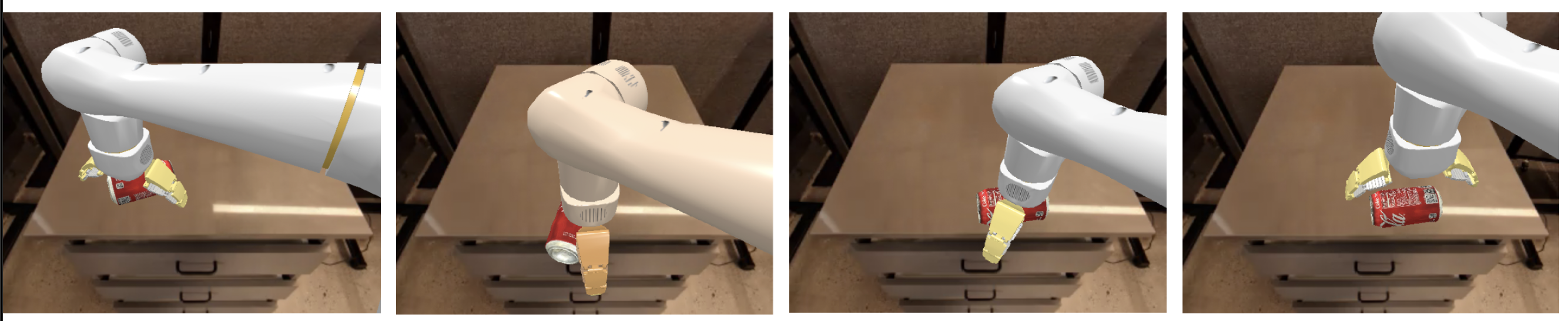}
    \caption{Examples that LMM fails to generate correct gripper pose under Euler representation.}
    \label{fig:euler_fail}
\end{figure}
\section{Complete Examples of Robot Manipulation Process as the Multi-round Conversations}
\label{appendix: examples_multi_round_conversations}
\begin{figure}[h]
    \centering
    \includegraphics[width=\linewidth]{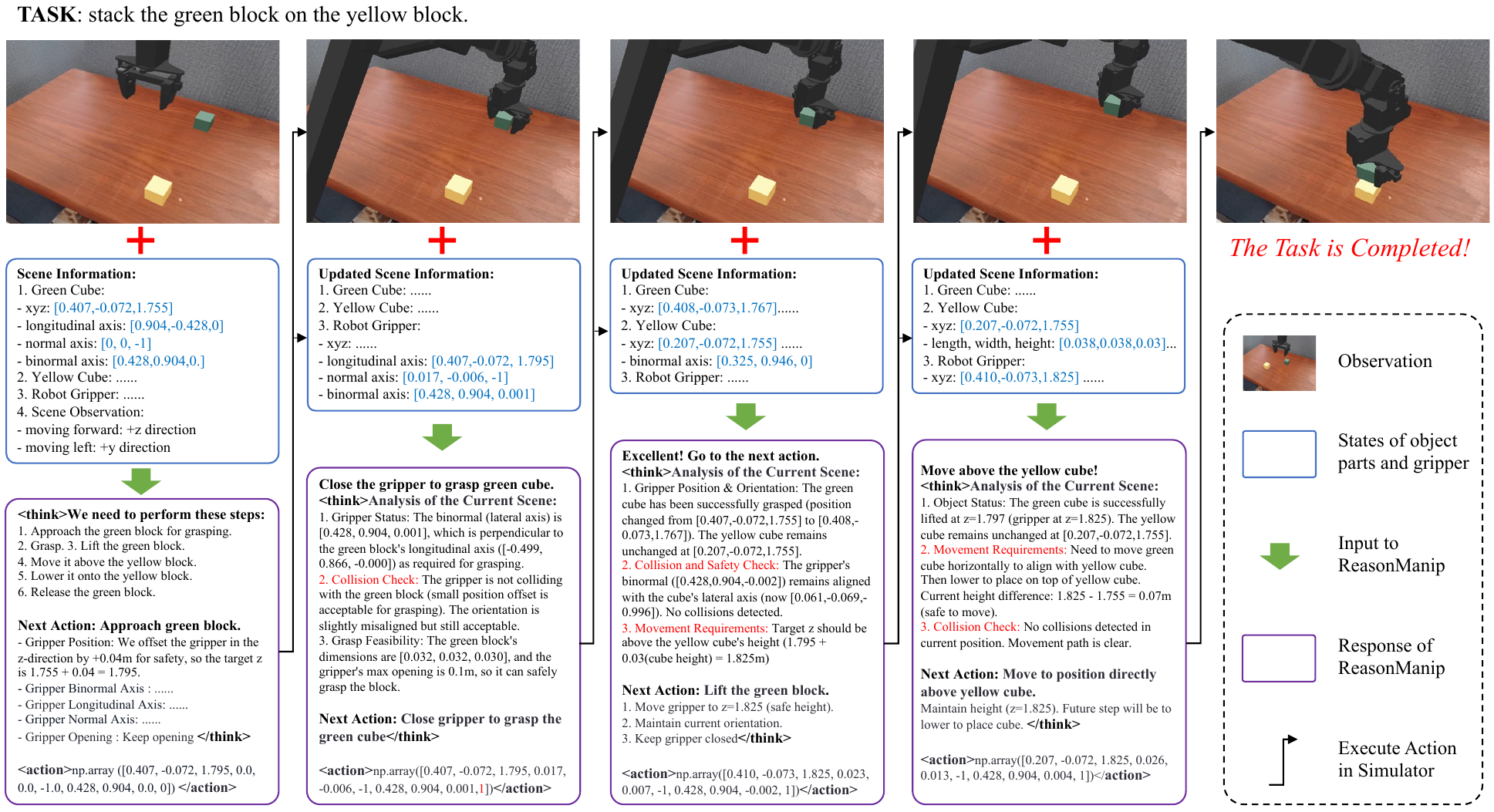}
    \caption{Summary of multi-round conversations for achieving "stack green cube on yellow cube".}
    \label{fig:round_summary}
\end{figure}

We provide a detailed example of the multi-round conversation for completing the task "Stack the green cube onto the yellow cube". First, we illustrate a summary version of the conversation in Fig.~\ref{fig:round_summary}.

\begin{figure}[h]
    \centering
    \includegraphics[width=0.95\linewidth]{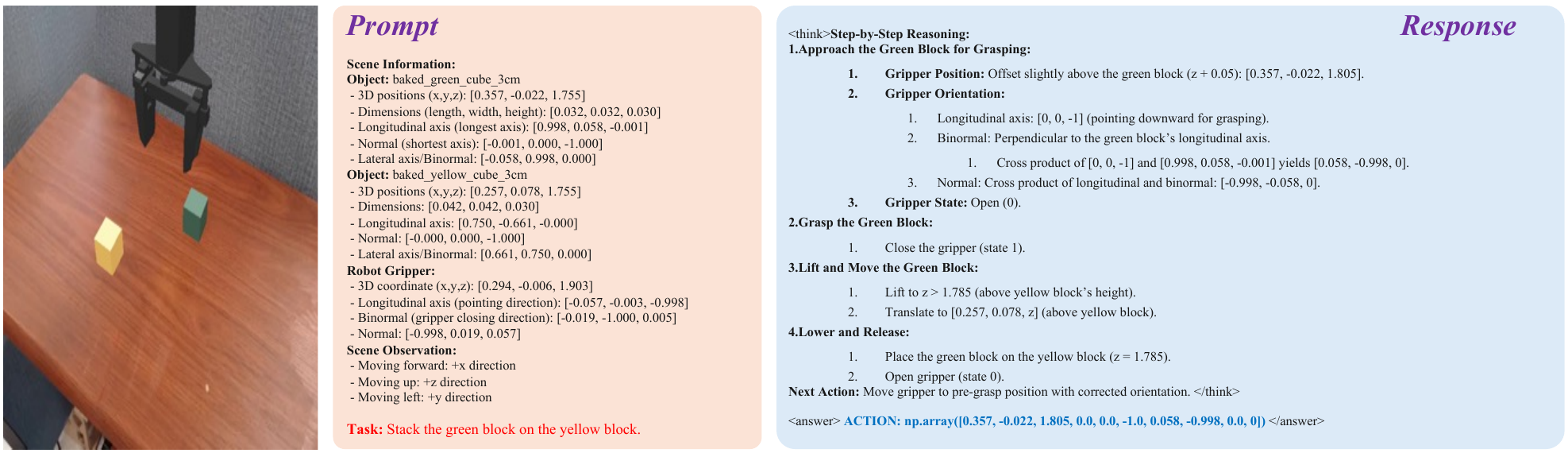}
    \caption{Conversation for "Stack the green cube onto the yellow cube" - Round 1}
    \label{fig:round_1}
\end{figure}

\begin{figure}
    \centering
    \includegraphics[width=0.95\linewidth]{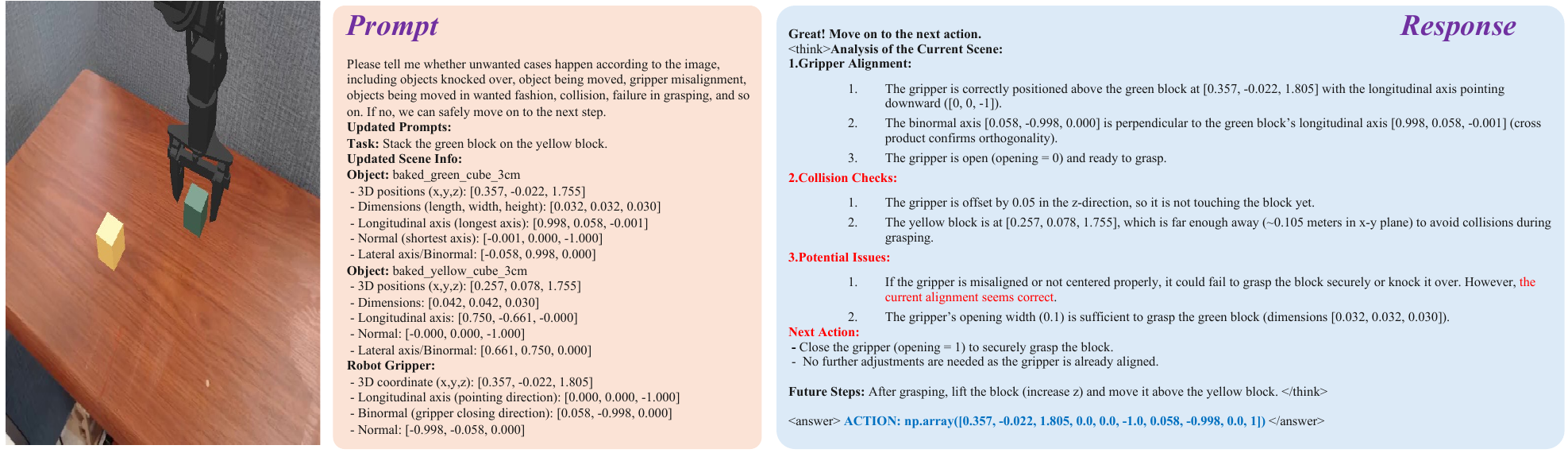}
    \caption{Conversation for "Stack the green cube onto the yellow cube" - Round 2}
    \label{fig:round_2}
\end{figure}

\begin{figure}[H]
    \centering
    \includegraphics[width=0.95\linewidth]{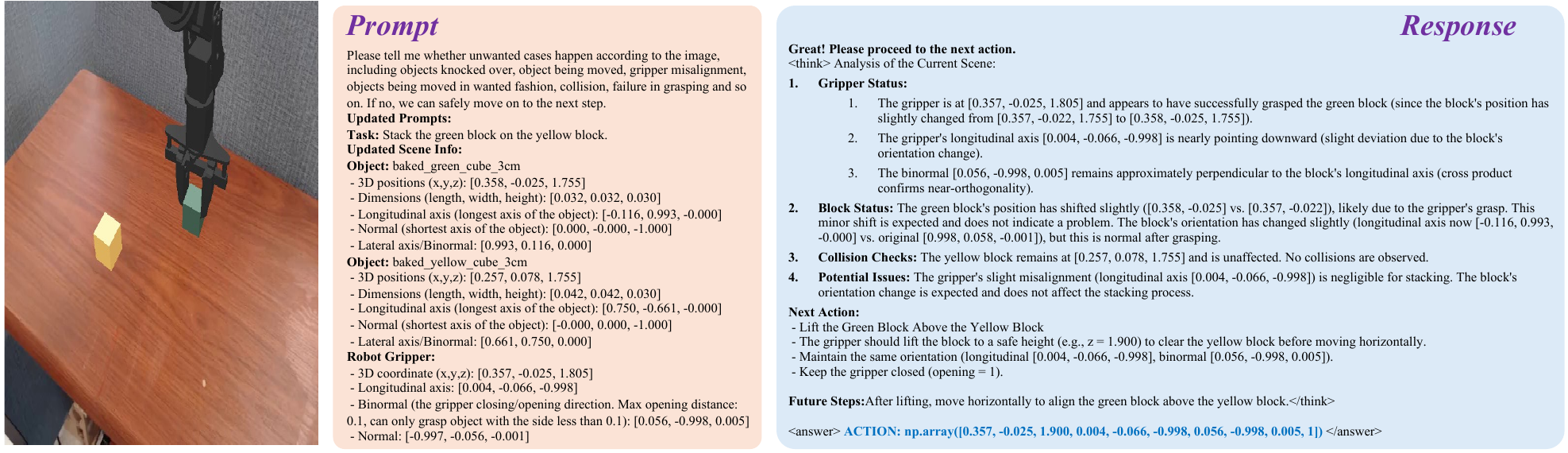}
    \caption{Conversation for "Stack the green cube onto the yellow cube" - Round 3}
    \label{fig:round_3}
\end{figure}

\begin{figure}[h]
    \centering
    \includegraphics[width=.95\linewidth]{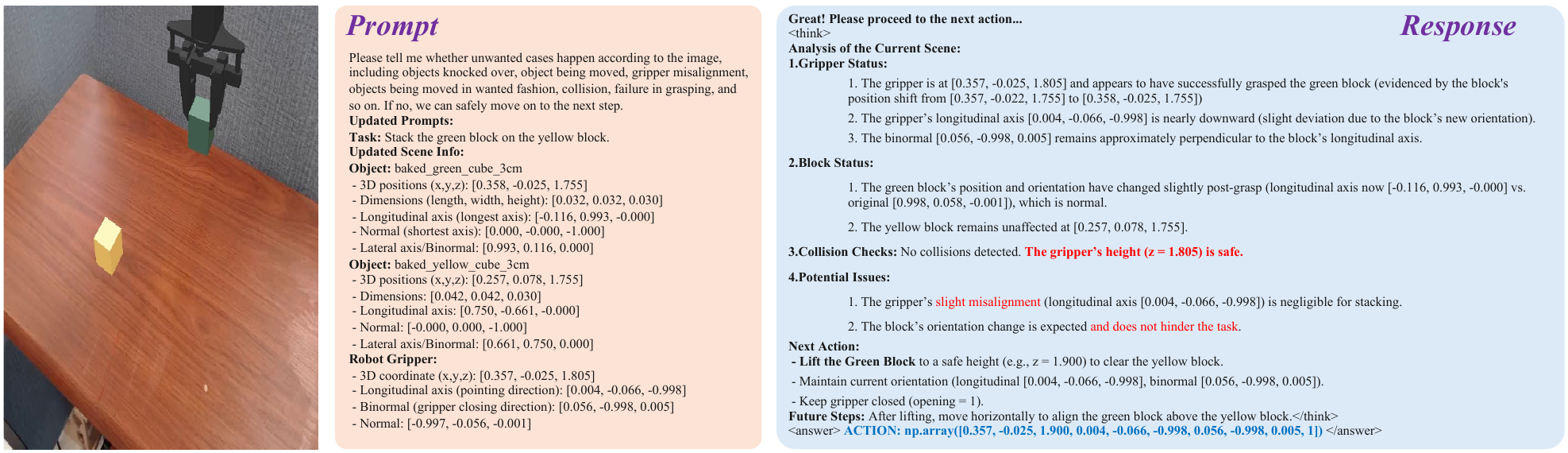}
    \caption{Conversation for "Stack the green cube onto the yellow cube" - Round 4}
    \label{fig:round_4}
\end{figure}

\begin{figure}[h]
    \centering
    \includegraphics[width=.95\linewidth]{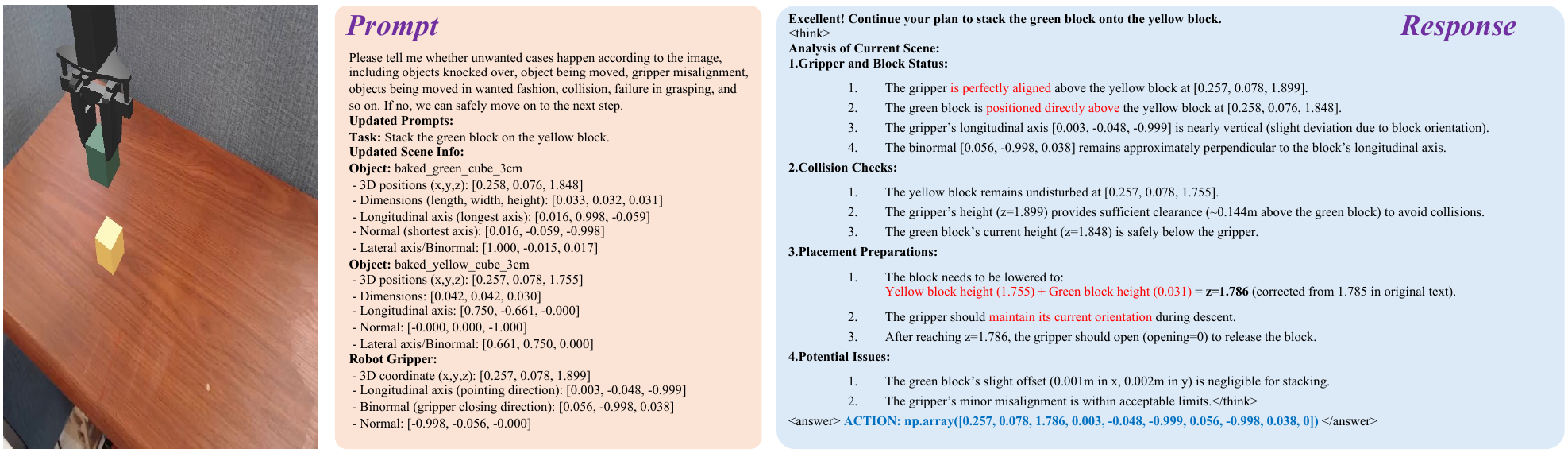}
    \caption{Conversation for "Stack the green cube onto the yellow cube" - Round 5}
    \label{fig:round_5}
\end{figure}

Next, we illustrate the complete output for this task in the following 5 figures: Fig.~\ref{fig:round_1}, Fig.~\ref{fig:round_2}, Fig.~\ref{fig:round_3}, Fig.~\ref{fig:round_4}, Fig.~\ref{fig:round_5} sequentially.


\section{Base Prompt}
\label{appendix:base_prompt}
Scene information:\{\}

Scene observation:\{\}

moving forward: +x direction
moving up: +z direction
moving left: +y coordinate

The task: \{\}.

Please, 1. Figure out the following steps to achieve the task. 2. For each step, solve for the gripper longitudinal axis, binormal, normal, and gripper locations for achieving the task. Notice that once grasped, the robot gripper is rigidly attached to the object, meaning the robot gripper's rotation and the grasped item's rotation are the same.
You should reason step by step in a chain-of-thought fashion. You can start your reasoning by analyzing the scenes. You should pay attention to potential collision between the gripper and objects. Add proper offset if appropriate. Please explain your answer in detail.

You should output a numpy array as the gripper target to move. It should be a standalone line and start this line strictly with `ACTION:`. It is in the shape of 10: np.array([gripper x location, gripper y location, gripper z location, 3 dimensions for gripper longitudinal axis, 3 dimensions for gripper lateral(binormal) axis, gripper opening/close (0 for opening, 1 for closing)]).

Since we don't know what will happen after the next move, you only need to predict the next action and discuss the future action given different situations after the next move.

To grasp something, the gripper's lateral axis should be perpendicular to the object's longitudinal axis. For grasping table-top objects, the gripper longitudinal should be np.array([0, 0, -1]) (pointing downwards).
gripper opening/closing should be a standalone step.

Please note only output your answer but also output your reasoning process: 

<think>Your thinking process here</think>

<answer>Your answer here</answer>

\section{Limitations and Future Works}
Our method typically fails when: (i) The object point cloud is not complete enough to derive accurate positions, axis representations, and scales. (ii) The manipulation trajectory is too curvy and complex to be straightforwardly solved by mathematical calculation (e.g., wipe the bowl with a rag). In the future, we plan to (i) use pose estimation or point cloud completions work to help mitigate missing point cloud. (ii) improve the control frequency for more complex task by automatically switching between system-2 reasoning and high-frequency direct action generation. (iii) build the \methodname based on larger and more intelligent LMM and expand it to wider tasks and scenarios.
\end{document}